\documentclass{article}
\usepackage{ijcai18}

\usepackage{times}
\usepackage{xcolor}
\usepackage{soul}
\usepackage[utf8]{inputenc}
\usepackage[small]{caption}

\usepackage{dblfloatfix}    
\usepackage{graphicx}
\usepackage{booktabs}
\usepackage[ruled, linesnumbered]{algorithm2e}
\usepackage{amsmath, amsfonts, dsfont}
\usepackage{amssymb, mathtools}
\usepackage{subcaption}
\usepackage{siunitx}

\newcommand{\R}{\mathbb{R}}

\newcommand{\EE}{\mathbb{E}}

\newcommand{\sset}{\mathcal{S}}
\newcommand{\unif}{\mathcal{U}}
\newcommand{\aset}{\mathcal{A}}

\newcommand{\trans}{\mathcal{P}}
\newcommand{\gset}{\mathcal{G}}

\title{Region Growing Curriculum Generation \\
for Reinforcement Learning}

\author{
Artem Molchanov$^1$, 
Karol Hausman$^1$, 
Stan Birchfield$^2$, 
Gaurav Sukhatme$^1$, 
\\ 
$^1$ University of Southern California \\
$^2$ Nvidia\\
}

\begin{document}

\maketitle

\begin{abstract}
Learning a policy capable of moving an agent between any two states in the environment is important for many robotics problems involving navigation and manipulation.
Due to the sparsity of rewards in such tasks, applying reinforcement learning in these scenarios can be challenging.
Common approaches for tackling this problem include reward engineering with auxiliary rewards, requiring domain-specific knowledge or changing the objective.

In this work, we introduce a method  based on region-growing that allows learning in an environment with any pair of initial and goal states. 
Our algorithm first learns how to move between nearby states and then increases the difficulty of the start-goal transitions as the agent's performance improves. 
This approach creates an efficient curriculum for learning the objective behavior of reaching any goal from any initial state.
In addition, we describe a method to adaptively adjust expansion of the growing region that allows automatic adjustment of the key exploration hyperparameter to environments with different requirements.
We evaluate our approach on a set of simulated navigation and manipulation tasks, where we demonstrate that our algorithm can efficiently learn a policy in the presence of sparse rewards.
\end{abstract}

\section{INTRODUCTION}
In recent years, deep reinforcement learning (Deep RL) has enjoyed success in many different applications, including playing Atari games~\cite{mnih2013playing}, controlling a humanoid robot to perform various manipulation tasks~\cite{chebotar16b,chebotar17a} and beating the world champion in Go~\cite{silver16alphago}. 
The success and wide range of use cases of RL algorithms is partly due to the very general description of the problem that RL aims to solve, i.e., to learn autonomous behaviors given a high-level specification of a task by interacting with the environment.
Such high-level specification is provided by a reward function, which must be sufficiently descriptive as well as easy to optimize for an RL algorithm to learn efficiently. 
These requirements make the design of the reward function challenging in practice, creating a bottleneck for even a wider set of applications for RL algorithms.

The problem of designing a reward function has been tackled in various ways.
These include: i) learning the reward function from human demonstrations in the field of inverse reinforcement learning (IRL)~\cite{levine11irl,abbeel04}, ii) initializing the reinforcement learning process with demonstrations in imitation learning~\cite{chebotar16b,kalakrishnan12}, and iii) creating reward shaping functions that aim to guide the RL process to high-reward regions~\cite{chebotar17a,popov17}.
Even though all of these methods have shown promising solutions to the problem of reward function design, they present other significant challenges such as the requirement of domain expertise or access to demonstration data.

Ideally, one would like to learn from a simple sparse binary reward that indicates completion of the task. 
Such a reward signal is natural for many goal-oriented tasks.
It allows significant reduction of engineering effort, and in some cases can be used to learn very complicated skills from human feedback, where design of the reward function is very hard~\cite{christiano17a}. 
Despite being attractive, such a reward function creates significant difficulties for learning. 
This is due to the fact that it is very unlikely for an agent to generate the exact sequence of actions leading to solving the task from random exploration~\cite{duan16a}.

Recent efforts focus on learning from such sparse reward signals by constructing a curriculum from a continuous set of tasks~\cite{Held17,florensa17a}. 
These methods exploit the simple intuition that tasks initialized closer to the goal should be easier to solve. 
Proximity to the goal is defined either explicitly~\cite{Held17} or through the number of random actions needed to reach the state from the goal~\cite{florensa17a}. 
Nevertheless, all of these methods have a common disadvantage: they are designed for either single-start or single-goal scenarios. 
In this paper, we address the situation in which the task contains both a continuous set of goals and a continuous set of initial conditions, thus broadening the applicability of our algorithm to a wide range of problems.
In addition, we introduce a method to adaptively adjust expansion of the growing region, eliminating manual tuning of a key exploration hyperparameter whose optimal value varies across different environments.




\section{RELATED WORK}

\textbf{Intrinsic motivation.}
Learning from sparse rewards is a long-standing goal in RL. 
The most established way of coping with such scenarios has been reward shaping~\cite{chebotar17a}, which requires extensive engineering and domain specific knowledge. 
To address this problem, various researchers proposed curiosity and intrinsic motivation~\cite{schmidhuber10,oudeyer07} as a more general way of guiding learning in the absence of the main reward. 
Intrinsic motivation is typically introduced in the form of auxiliary rewards or loss components incentivizing exploration, that are not connected to the main objective. Such incentives could be based on counting visited states and/or maintaining a state-visitation density model~\cite{bellemare16a,ostrovski17a,martin17}, prediction error~\cite{stadie15}, prediction error-improvement of the learned model~\cite{lopes12}, predictive model uncertainty~\cite{houthooft16}, neuro-correlation~\cite{schossau16} or learning auxiliary tasks~\cite{jaderberg16}.
Despite a vast variety of approaches, many curiosity-inspired methods are prone to creating additional local minima in the learned objective function.

\textbf{Curriculum learning.}
Another approach to learning in the presence of sparse rewards is to construct a \emph{curriculum} of the task instances to ease the learning process. 
In this case, the agent initially learns from easy scenarios, where the chance of acquiring positive reward is relatively high, and the difficulty of the presented tasks is gradually increased until the final task is learned. 
The main advantage of such an approach is that the agent learns on the final objective directly, and thus avoids the problems of curiosity-driven methods. 
Traditionally, curriculum design has been explored from the perspective of manually engineered schedules in both supervised tasks~\cite{zaremba14,bengio15a} and reinforcement learning scenarios~\cite{wu17a,hees17}. 
More recently, there have also been multiple approaches for automated curriculum generation for RL. 
\cite{svetlik17} create curriculum in the form of an acyclic graph based on a \textit{transfer potential} metric,~\cite{sharma17a} explored task sampling based on their current performance, and~\cite{matiisen17} utilized task performance improvement as a basis for task sampling. 
All of these approaches, as opposed to our method, are designed to perform well only in a discrete set of tasks with dense rewards. 

Another related approach is presented in the recent work by \cite{Sukhbaatar17} based on the idea of self play between two agents. 
The first agent plays the role of a teacher that sets the tasks for the second agent, who plays the role of a student who tries to repeat the teacher's actions or reverse the environment to its original state. 
As mentioned by the authors and confirmed in~\cite{florensa17a}, the asymmetric structure of this method often leads to a biased exploration resulting in the teacher and the student becoming stuck in a small subspace of the task. 
Our method avoids such situations by using random exploration to expand the set of goals and the initial conditions to the appropriate level of difficulty. 

Another related piece of work is that of~\cite{Held17}, who consider the problem of generating multiple goals of the appropriate level of difficulty using generative adversarial networks (GANs)~\cite{goodfellow14}. 
Their approach is designed to learn a goal distribution and, thus, in its straightforward form, cannot learn to generalize to multiple initial conditions. 
In addition, since their approach contains a learned generative model, it tends to struggle when the dimensionality of the task representation is large and the number of examples is very limited, which is usually the case for robotics. 
We address this problem by generating tasks through the interaction with the environment.

The approach most related to ours is the concurrent work of~\cite{florensa17a}. 
We exploit similar core principles and assumptions, i.e., we utilize Brownian motion for growing the current task region and generate curriculum through reverse exploration of new tasks. 
We extend this approach to multi-goal and multi-start scenarios with infinitely many start-goal pairs, and present results in environments with sparse rewards. 
In addition, we address the question of controlling expansion of the growing region. 
Our algorithm adaptively changes the key exploration hyperparameter for environments with significantly different optimal settings.
These contributions lead to improved resampling efficiency and eliminate the need of expensive hyperparameter tuning.

\section{Background}
\label{sec:problem_definition}

We consider a reinforcement learning problem where an agent is represented by a global policy that aims to reach any goal in an environment. 
This section introduces a formal definition of the problem and our framework.

\subsection{Markov decision process}
\label{sec:mdp}
We consider a discrete-time, finite-horizon Markov decision process (MDP) defined by a tuple ${M = (\sset, \gset, \aset, \trans, r, \rho_0, T)}$, in which 
$\sset$ is the agent state set, 
$\aset$ is the action set, 
${\trans: \sset \times \aset \rightarrow \mathbb{R}^{n}}$ is the transition probability distribution, 
${r: \sset \times \gset \times \aset \rightarrow \R}$ is a bounded reward function dependent on the goal state, 
where $\gset$ represents the goal set; ${\rho_0: \sset \to \mathbb{R}^{n}}$ is the initial state distribution, 
and $T \in \mathbb{N}$ is the time horizon. 
Our aim is to learn a stochastic policy ${\pi_{\theta}: \sset \times \aset \times \gset \to \mathbb{R}^{m}}$ parameterized by $\theta$. 
We would like to point out that in order to communicate the goal to the agent, our formulation requires the policy to be conditioned on the goal $g$ specified by the environment, i.e. ${\pi_{\theta} = \pi(a_t | s_t, g)}$. 
The objective is to maximize the expected return, ${\eta_{\rho_0}(\pi_\theta) = \EE_{s_0\sim\rho_0, g\sim\rho_{g}} R(\pi_\theta, s_0, g)}$ with the expected reward starting at $s_0$ being $R(\pi_\theta, s_0, g) := \EE_{\tau|s_0}[ \sum_{t=0}^T r(s_t, a_t, g) ]$, where ${\tau = (s_0, a_0, \ldots)}$ denotes the trajectory generated by executing actions ${a_t \sim \pi_\theta(a_t|s_t, g)}$ sampled from the policy under environment dynamics ${s_{t+1} \sim \trans(s_{t+1} | s_t, a_t)}$. 

\subsection{Goal-dependent sparse reward function}
\label{sec:reward_func}
In this work, we consider the problem of reaching any goal state $g \sim \unif(\gset)$ in the environment from any initial state $\rho_0 \sim \unif(\sset^{0})$, where $\unif$ denotes a uniform distribution. 
For this purpose, we define a sparse binary reward function dependent on the goal:
\begin{equation}
\label{eq:reward} 
r(s_t, a_t, g) = \mathds{1}\{s_{t} \in S^g\} \,,
\end{equation}
where $S^g \subset \sset$ is a set of states corresponding to the goal $g$. 
We note that although the binary reward function in Eq.~\eqref{eq:reward} is typically defined through some distance metric $\epsilon$, our learning algorithm does not explicitly utilize this metric. 

\subsection{Assumptions}
\label{sec:assumptions}
In this work we exploit several assumptions:

\textbf{Assumption 1} \textit{The agent can be initialized at an arbitrary state $s \in \sset$.} 
This assumption is a common requirement for many algorithms~\cite{florensa17a,kearns02} in the RL setting, especially those that exploit uniform initialization to generalize to multiple initial states.

\textbf{Assumption 2} \textit{At least one initial state is provided to the algorithm, which we call a seed state.}  

\textbf{Assumption 3} \textit{For every state $s \in \sset$, there exists a function $g = f_{g}(s)$ that maps any state in the environment to the corresponding goal representation.} 
This assumption is required since, in our algorithm, states encountered by the agent should be converted to the corresponding goal representations. 

\textbf{Assumption 4} \textit{For any pair of states $s_1, s_2 \in \sset$ there exists a trajectory that moves the agent from $s_1$ to $s_2$.} In other words, the agent can reach any state from any other state.

We note that although we explicitly introduced Assumption 4, it does not prevent our algorithm from being applied to a wider set of tasks where some states might not be mutually reachable. 
For example, if isolated or irreversible pairs of states exist, the algorithm nevertheless applies to all the reachable states, which depends on the initial state provided. 


\section{Approach}
The main difficulty of training an RL agent in a sparse reward setting arises from the fact that it is unlikely for the agent to accomplish the task using random exploration if the initial state is far from the goal state.
In this work, we take advantage of the intuition also exploited by~\cite{florensa17a} that the agent has a higher chance of success if the goal is located in  close proximity to the initial state.
In particular, if one could initialize learning by generating goal states that are close to the initial states, then the initial learning stages should progress much faster.

Since it can be highly nontrivial to engineer a correct distance metric directly in the observation space, we define the proximity of points by the number of actions it takes to reach one point from another.


Taking this into consideration, we propose the idea of gradually-growing \textit{reachability regions} for generating a curriculum in a multi-goal setting. 
Our algorithm consists of two agents: a sampler and a learner. 
The sampler uses short chains of random actions to arrive at a state that is then added to the currently-explored set of points, which we refer to as the reachability region.
This region is defined as the area where the learner has already mastered transitions between all pairs of points.
As learning progresses, the sampler removes already-learned states from the reachability region and adds new points that have not been explored yet.
This generates a natural curriculum for learning a global reaching policy, i.e., a policy capable of moving the agent between any two states in the environment.
In the following, we first discuss the sampler and then describe the learner.

\subsection{Filtering states}
\label{sec:filter_states}
In the first part of the sampling algorithm, we focus on a criterion that indicates whether a particular set of states has been mastered.
In order to select which states have already been mastered, we retain statistics of rewards received by the agent on every state within the current reachability region. 
We choose to follow a simple approach in which we only retain statistics of the points in the role of starting states, as opposed to retaining statistics on start-goal pairs.
We define thresholds $R_{min}$ and $R_{max}$ that prevent states from being too hard or too easy, respectively.
We refer to the set of all states in the current reachability region as $s$.
We keep a history of rewards in a vector $r$ and associate them with start states.
If the average reward for a state in $r$ does not exceed the $R_{min}$ and $R_{max}$ thresholds we use the state for further resampling.
This behavior is implemented in a helper function \texttt{FilterStates} that takes $s$, $r$, $R_{min}$, and $R_{max}$ as input and returns the retained set of states as $s$.



\subsection{Adaptive state resampling}
\label{sec:resampling}
As previously mentioned, we define the proximity of the points through the action space, i.e., points are close to each other if they are reachable via short random trajectories. 
We use Brownian motion to sample new states to grow the region of learned state-goal pairs. 


A major challenge of this approach is the selection of the variance for exploration.
Poorly selected variance can result in either a very spread set of points that are hard to learn from, or a set of points that are too easily mastered, which in both cases results in a slow learning progress of the RL algorithm.
We adjust the sampling variance $\sigma$ dynamically using a method that is inspired by the integral part of a PID controller.
Our approach adjusts the variance such that average reward in the current iteration ($r_{avg}$) is close to a user-provided target reward ($R_{pref}$).
In particular, every time before resampling, we update the sampling variance ($\sigma$) according to the following procedure:
\begin{align}
    \delta^\sigma &\leftarrow \text{Clip}( k_{\sigma} * (r_{avg} - R_{pref}), ~-\delta^{\sigma}_{max},  ~\delta^{\sigma}_{max}) \nonumber \\
    \sigma~  &\leftarrow \text{Clip}(\sigma + \delta^\sigma, ~\sigma_{min}, ~\sigma_{max})
    \label{eq:var_adapt}
\end{align}
where 
$\text{Clip}(x, \alpha, \beta) \triangleq \min(\max(x, \alpha), \beta)$,
$k_{\sigma}$ is the control coefficient,
$\delta^{\sigma}_{max}$ is the maximum change of variance, and $\sigma_{min/max}$ are the variance limits.
Thus, if the success ratio systematically exceeds the preferred value, our method increases the variance, promoting faster exploration and vice versa.

We encode Eq.~\eqref{eq:var_adapt} in the helper function \texttt{UpdateVariance} that takes $\sigma$ and $r_{avg}$ as inputs and returns the new sampling variance.
Resampling a set of new states is implemented in the helper function \texttt{ResampleStates} that takes the current set of states $s$, the set of old mastered states $s_{old}$, and the variance $\sigma$ as inputs and returns the the new set of states.
Resampling is carried out in two stages. 
First, we create an oversampled set of states by performing Brownian-motion rollouts, which we refer to as sampling rollouts.
We use random actions generated by the sampler agent using $\mathcal{N}(0, \sigma^2)$ and collect states visited by the agent.

Each of these rollouts is initialized at one of the states from the growing oversampled set.
This set is initialized with the states retained in $s$ after filtering. 
At the second stage, we sample $N_{new}$ states uniformly from the oversampled set and add them to $N_{old}$ states sampled uniformly from $s_{old}$ to form the new current set of states.

\subsection{Policy training}
\label{sec:policy_train}

\begin{algorithm}
\SetAlgoLined
\SetKwInOut{Input}{Input}
\SetKwInOut{Output}{Output}

\Input{$s_{seed}$: seed state, $N$: iterations, $K$: sampling period, $\pi_1$: initial policy, $\sigma$: initial sampling variance}
\Output{$\pi_{N+1}$: policy}

 $s_{old}, s, r \leftarrow \{s_{seed}\}, \{s_{seed}\}, [1]$ \label{alg1:init}

 $s$ $\leftarrow$ ResampleStates($s$, $s_{old}$, $\sigma$)
 
 \For{$i \leftarrow \ 1$ \KwTo $N$}{
 
    
    \# Every $K$'th iteration
    
    \If{$i\;\mathrm{mod}\;K = 0$}{ \label{alg1:samplerBegin}
      \# See Eq.~\eqref{eq:var_adapt}
    
      $\sigma$ $\leftarrow$ UpdateVariance($\sigma$, $r_{avg}$) \label{alg1:updateVariance}
    
      \# See Sec.~\ref{sec:filter_states}
    
      $s$ $\leftarrow$ FilterStates($s$, $r$, $R_{min}$, $R_{max}$) \label{alg1:filterStates}
      
      $s_{old} \leftarrow s_{old} \cup s$
       
      \# See Sec.~\ref{sec:resampling}
       
      $s$ $\leftarrow$ ResampleStates($s$, $s_{old}$, $\sigma$) \label{alg1:resampleStates}

    } \label{alg1:samplerEnd}
    
    $s_{train}$, $g_{train}$, $r$, $r_{avg}$ $\leftarrow$ Rollouts($\pi_{i}$, $s$, $s_{old}$) \label{alg1:learnerBegin}

    $\pi_{i+1}$ $\leftarrow$ UpdatePolicy($\pi_{i}$, $s_{train}$, $g_{train}$, $r$) \label{alg1:learnerEnd}

 }
 \caption{Policy Training}
 \label{alg:policy_train}
\end{algorithm}

Algorithm~\ref{alg:policy_train} describes the policy training procedure including both the sampler and the learner agents.
The sampler agent updates the reachability region (lines \ref{alg1:samplerBegin} -- \ref{alg1:samplerEnd}), while the learner follows a its own learning strategy (lines \ref{alg1:learnerBegin} -- \ref{alg1:learnerEnd}).

Our method starts by initializing the current state set $s$, the corresponding vector of history of rewards $r$ and the pool of the previously learned states $s_{old}$ (line \ref{alg1:init}).


The sampler uses a fixed update period $K$ (line \ref{alg1:samplerBegin}) to adjust the variance according to Eq.~\eqref{eq:var_adapt} (line \ref{alg1:updateVariance}) and proceeds to the filtering stage to find good states to propagate from (line \ref{alg1:filterStates}).
Once the filtering is finished, the sampler resamples a new set of states using Brownian motion (line \ref{alg1:resampleStates}).

The learner performs policy rollouts in every iteration (line \ref{alg1:learnerBegin}) using the helper function \texttt{Rollouts}.
This function follows a special start-goal pair sampling strategy. Start states for the rollouts are sampled uniformly from the current state set $s$, whereas the goals are sampled from either $s$ (with probability $P_{new}$) or $s_{old}$ (with probability $1 - P_{new}$).
Once the batch of samples used for the user-chosen RL algorithm is accumulated, we update the policy (line \ref{alg1:learnerEnd}).

Our approach is agnostic to the choice of agent optimization method; we only require that this method provides the \texttt{UpdatePolicy} function.
In our experiments we use  TRPO~\cite{schulmanTRPO} as one of the most robust RL algorithms with an implementation available online.

\section{Experiments}



We implemented this approach in Python and applied it to two representative environments.
We show empirically that this technique successfully trains agents in our multi-goal scenarios.
Furthermore, we demonstrate that our dynamic variance selection is less sensitive to hyperparameters than other alternatives.
In all of our experiments, we use the following parameters across all environments:
$R_{max} = 0.9$, $R_{min} = 0.3$, $K=5$, $N_{new} = 135$, $N_{old} = 65$, $P_{new}=0.6$, $R_{pref}=0.7$, $k_{\sigma}=2.0$, $\delta^{\sigma}_{max} = 0.5$, $\sigma_{min} = 0.1$, $\sigma_{max} = 1.0$. 

\begin{figure}
\includegraphics[width=0.225\textwidth]{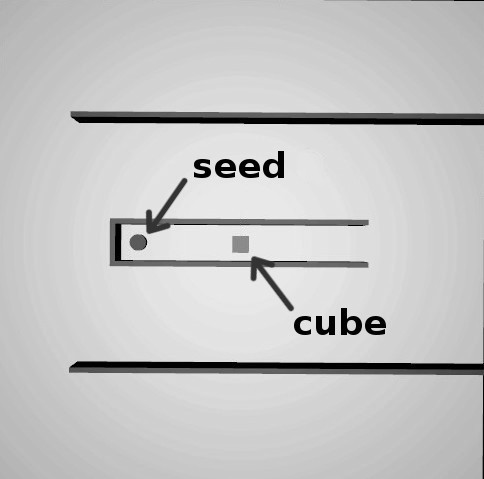}
\includegraphics[width=0.225\textwidth]{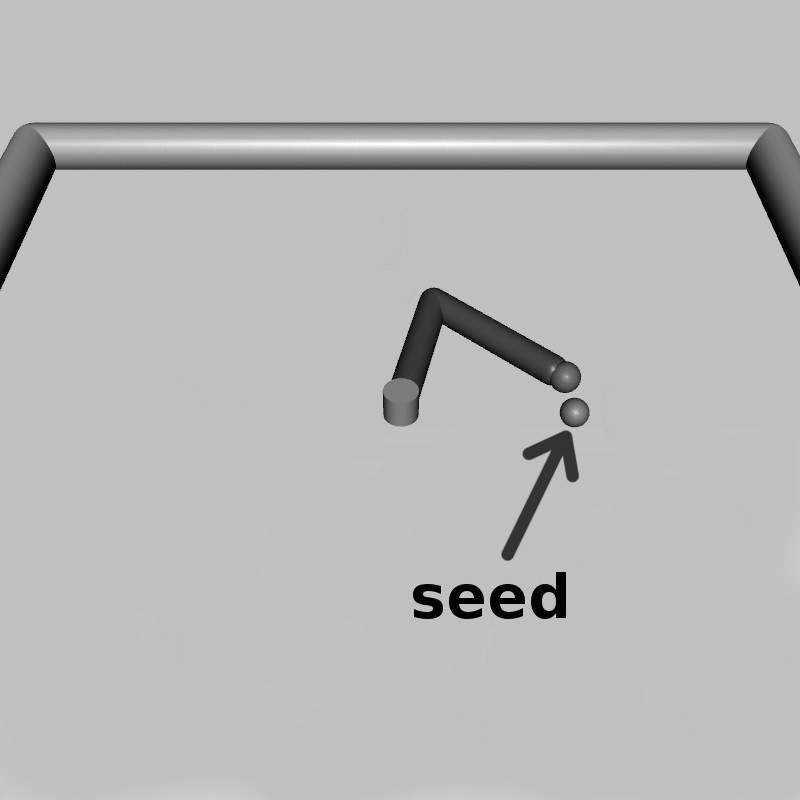}
\caption{Environments with seed states used in our experiments. Left: Maze environment. The square represents the cube that the agent has to push to a goal state. Black lines represent the walls of the maze. Right: SparseReacher environment. The two-link manipulator has to touch the goal marker. 
}
\label{fig:envs}
\end{figure}

\begin{figure*}[t]
\begin{subfigure}[b]{0.24\textwidth}
\setlength{\fboxsep}{0pt}
\fbox{\includegraphics[width=\textwidth]{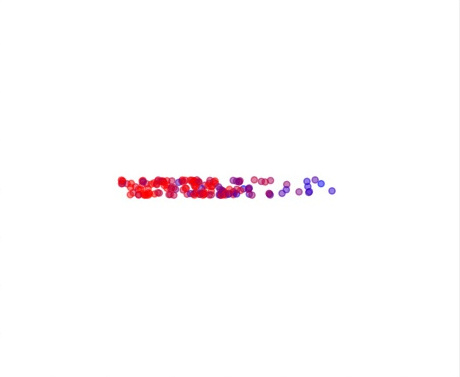}}
\caption{$i=5$}
\label{maze_itr0}
\end{subfigure}
\hfill
\begin{subfigure}[b]{0.24\textwidth}
\setlength{\fboxsep}{0pt}
\fbox{\includegraphics[width=\textwidth]{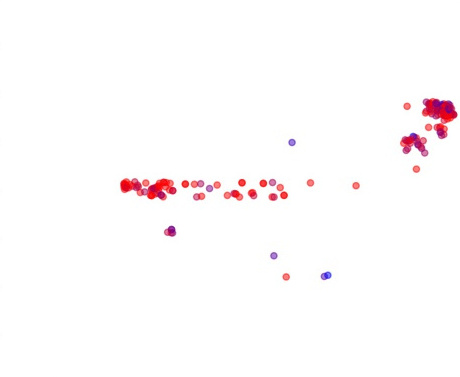}}
\caption{$i=125$}
\label{maze_itr1}
\end{subfigure}
\hfill
\begin{subfigure}[b]{0.24\textwidth}
\setlength{\fboxsep}{0pt}
\fbox{\includegraphics[width=\textwidth]{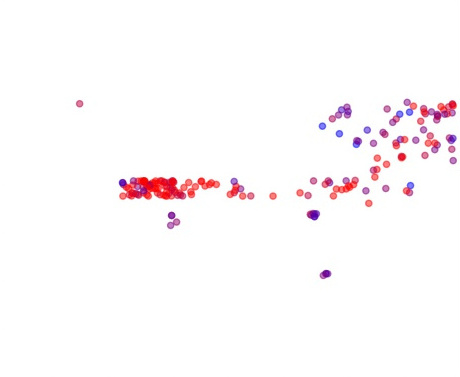}}
\caption{$i=135$}
\label{maze_itr2}
\end{subfigure}
\hfill
\begin{subfigure}[b]{0.24\textwidth}
\setlength{\fboxsep}{0pt}
\fbox{\includegraphics[width=\textwidth]{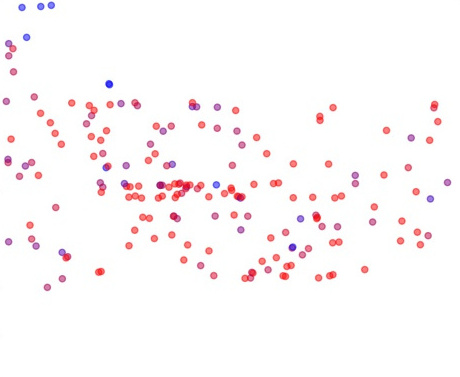}}
\caption{$i=450$}
\label{maze_itr3}
\end{subfigure}
\caption{Illustration of state propagation for the maze multi-goal environment. Circles represent the current states in the reachability region. 
Images are ordered from left to right in the order of learning progress. 
The leftmost image depicts the beginning of training and the rightmost image shows the state set at the end of training. The middle two plots show the phenomenon of state clustering. 
Colors encode average reward associated with the states, where red refers to high reward and blue to low reward.}
\vspace{-10pt}
\label{fig:maze_propagation}
\end{figure*}

\subsection{Environments}
The \textit{SparseReacher} is an environment with a two-link manipulator based on the Reacher-v0 environment from OpenAI Gym~\cite{gym}. 
We use it in a sparse reward setting: the agent receives a positive binary reward only when it touches the goal marker.
This corresponds to the situation where the robot's end effector is not further than $\SI{2}{cm}$ from the center of the goal marker.
In addition, the Cartesian velocities of the robot must be lower than $\SI{0.2}{m/s}$.
The episode ends as soon as the positive reward is acquired.
As we observed in our experiments, such sparse reward makes this environment significantly more challenging, especially when the goal is to learn a policy that can reach any point in the robot's workspace.

The goal in the \textit{Maze} environment is to bring a cube of size $h_{cube}$ to a goal location.
The agent receives a reward only if the center of the cube lies still within an {$\epsilon$-radius} of the goal location. 
The episode ends as soon as the positive reward is acquired.
We define a variable time step in the environment that is dependent on the time it takes for the cube to settle after a force is applied. 
The table is constrained by the size $h_{table}$ and surrounded by walls, such that the cube cannot fall off the table. 
This environment has continuous action space that consists of two components of the force $F_{x},F_{y}$ applied to the center of the cube, parallel to the table plane. 
Observations contain a 7-dimensional cube pose where the rotation is encoded as quaternion. 
We define a goal representation as a simple 2d position on the table. 

This environment is challenging due to several aspects. 
First, the search space increases with $h^2_{table}$, thus, the probability to encounter the target by chance is very small. 
Second, the cube has relatively complex dynamics compared to a simple point mass: it can be pushed or rolled depending on the direction and amount of force applied, and it exhibits a complex behavior when it comes into a contact with the wall. 
Third, the cube cannot simply roll over the goal to acquire a positive reward---it must stop at the precise location of the goal. 

Both environments are shown in Fig.~\ref{fig:envs}.
For each environment, we select a single \textit{seed} state to expand the growing region.
For the Maze environment, we explicitly pick the most challenging scenario of the seed state located at the end of the central corridor since the policy has to learn how to precisely navigate inside of the narrow corridor entrance. Both environments can be naturally used in both single- and multi-goal settings. 
We also note that in every training scenario, in addition to the sparse reward, we add a very small negative reward for every time step to promote shorter episodes.


    

\subsection{Reachability regions}

\begin{figure*}[t]
\begin{subfigure}[b]{0.49\textwidth}
\centering
\includegraphics[width=\textwidth]{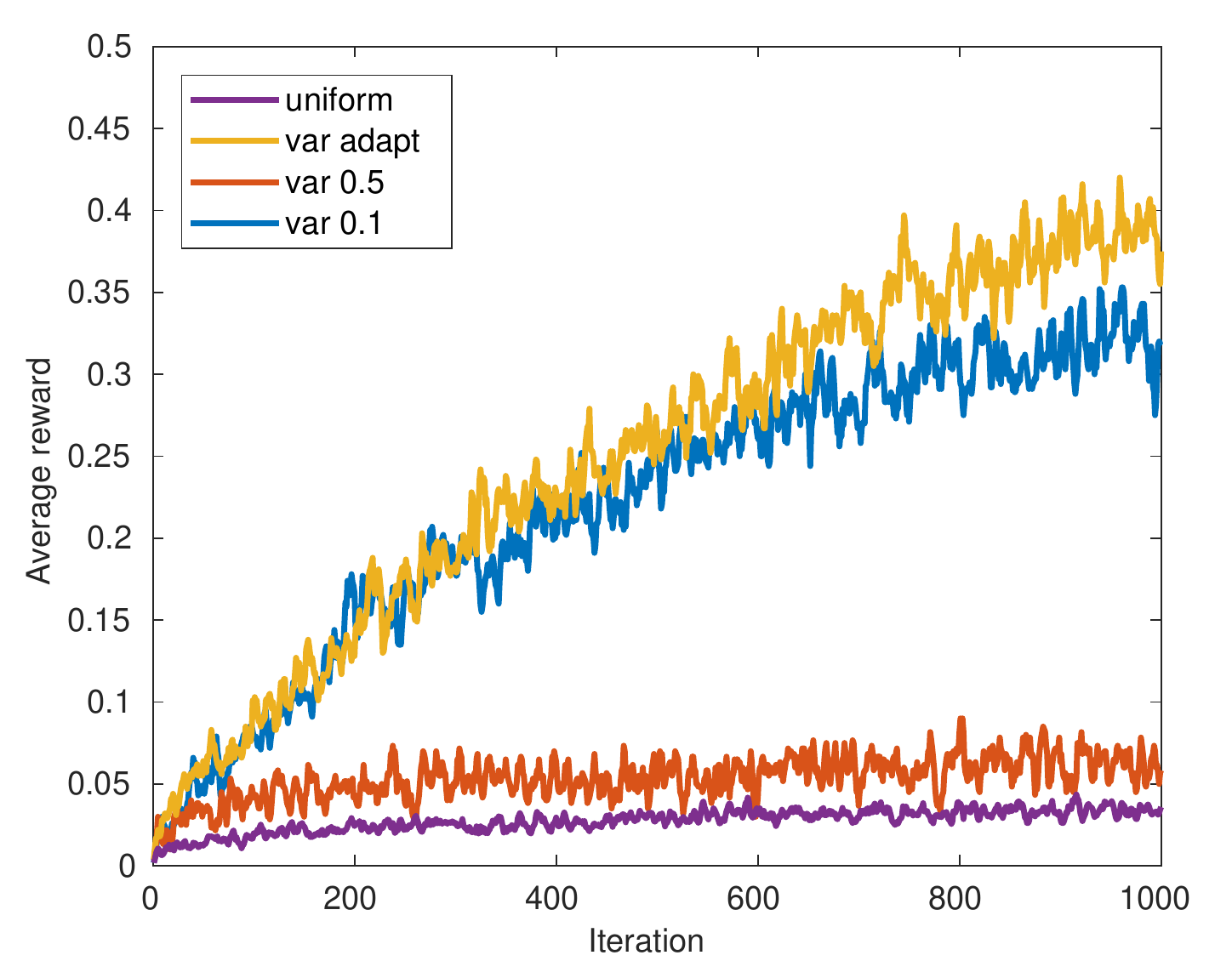}
\caption{SparseReacher environment.}
\label{reacher_variances}
\end{subfigure}
\hfill
\begin{subfigure}[b]{0.49\textwidth}
\centering
\includegraphics[width=\textwidth]{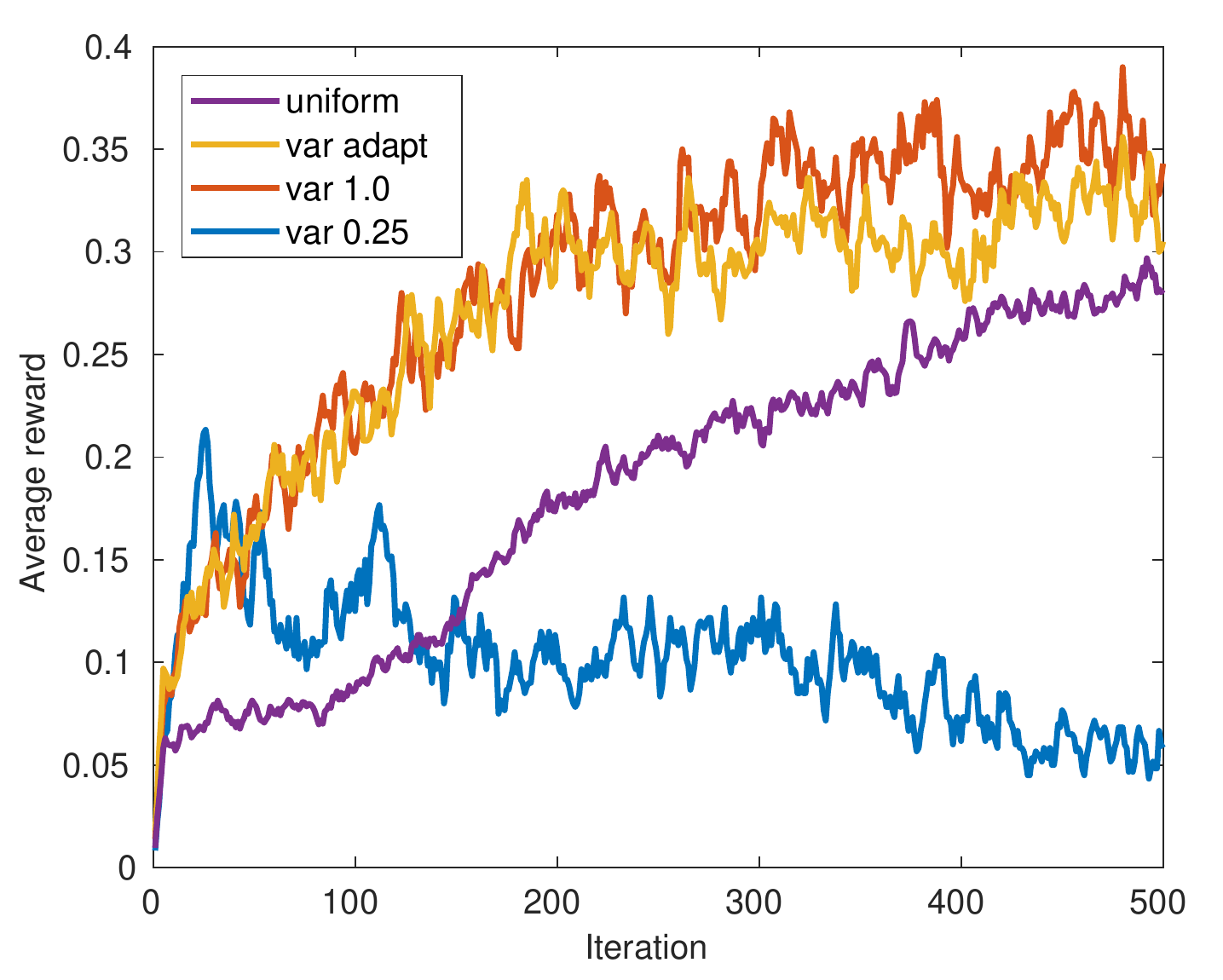}
\caption{Maze environment.}
\label{maze_variances}
\end{subfigure}
\caption{Reward for different algorithm variants for the multi-goal case. The data is averaged over 6 executions. ``Uniform'' refers to uniform random sampling of the start and goal states with no reachability region. ``var $x$'' uses the reachability regions for the sampler agents, but uses a constant $\sigma=x$ for action selection. ``var adapt'' is our full algorithm using the reachability regions and adaptive $\sigma$.}
\label{expResults}
\vspace{-10pt}
\end{figure*}

Fig.~\ref{fig:maze_propagation} demonstrates how the region expansion proceeds during learning in the maze environment.
In particular, it shows an interesting phenomenon associated with variance adaptation that we refer to as region clustering.
During expansion, if the new set of points was selected too aggressively, our algorithm responds by decreasing the variance of the region expansion.
Since this event by definition happens due to a poor performance (see Eq.~\eqref{eq:var_adapt}), there will be very few available states to sample from.
Thus, the algorithm forms a cluster of newly resampled states located around a few states that passed through the filtering stage (see Fig.~\ref{maze_itr1}).
Later, as the learner agent improves, those clusters grow and connect, forming a single region which is illustrated in Fig.~\ref{maze_itr2}.
Such behavior helps the learner to create new growing regions in isolated areas.

\subsection{SparseReacher}

Our results for the multi-goal version of the SparseReacher environment are shown in Fig.~\ref{reacher_variances}.
We execute the learning process several times and provide the average reward for each iteration over six executions.
We test our algorithm with two constant resampling variances ($\sigma=0.1$ and $\sigma=0.5$) and with our adaptive variance.
We also provide results for the case that does not use a reachability region, but instead samples start and goal states uniformly over the environments.

The environment is conservative and requires small exploration variances; we found that a constant variance of $\sigma=0.1$ performs much better than a variance of $\sigma > 0.5$.
Our adaptive variance selection achieves a slightly higher average reward than the best hand-tuned constant variance.
The simple uniform state sampling performs as well as our reachability approach when a bad constant variance is applied.

\subsection{Maze}

The experimental results for the multi-goal version of the Maze environments are shown in Fig.~\ref{maze_variances}.
As before, we provide the average reward for each iteration over six executions.

This environment requires more exploration than the SparseReacher environment; we found that when using a constant variance a value of $\sigma=1.0$ the agent performs best, while $\sigma < 0.25$ results in a very poor learning performance.
The reward of our adaptive variance selection is comparable to the best hand-tuned constant variance.

Uniform state-goal sampling performed surprisingly well, but as we can see our approach clearly indicates the benefits of generating a curriculum for learning.
We would also like to note that uniform sampling exhibits the best of its capabilities to learn transitions in this environment, whereas we selected the worst seed state for our algorithm specifically to emphasize benefits of our dynamic reachability region.

\subsection{Hyperparameter adaptation}

The environments that we selected are representative in the spectrum of requirements for growing region expansion. 
The multi-goal version of the SparseReacher environment is more conservative and requires small exploration variances, whereas the Maze environment benefits from aggressive exploration, and hence high variances perform well.
For example, Fig.~\ref{reacher_variances} shows that, under constant variance, the learner completely fails to improve when the variance is set to a high value.
On the other hand, Fig.~\ref{maze_variances} shows the opposite for the Maze, where the optimal variance value is close to the maximum value. 
Our adaptive variance approach performs similar to the optimal constant variance.
Given that we have the same set of exploration hyperparameters for both environments, our approach eliminates the need to tune the key hyperparameter of the region growing curriculum learning method.

\begin{figure}
\centering
\includegraphics[width=0.40\textwidth]{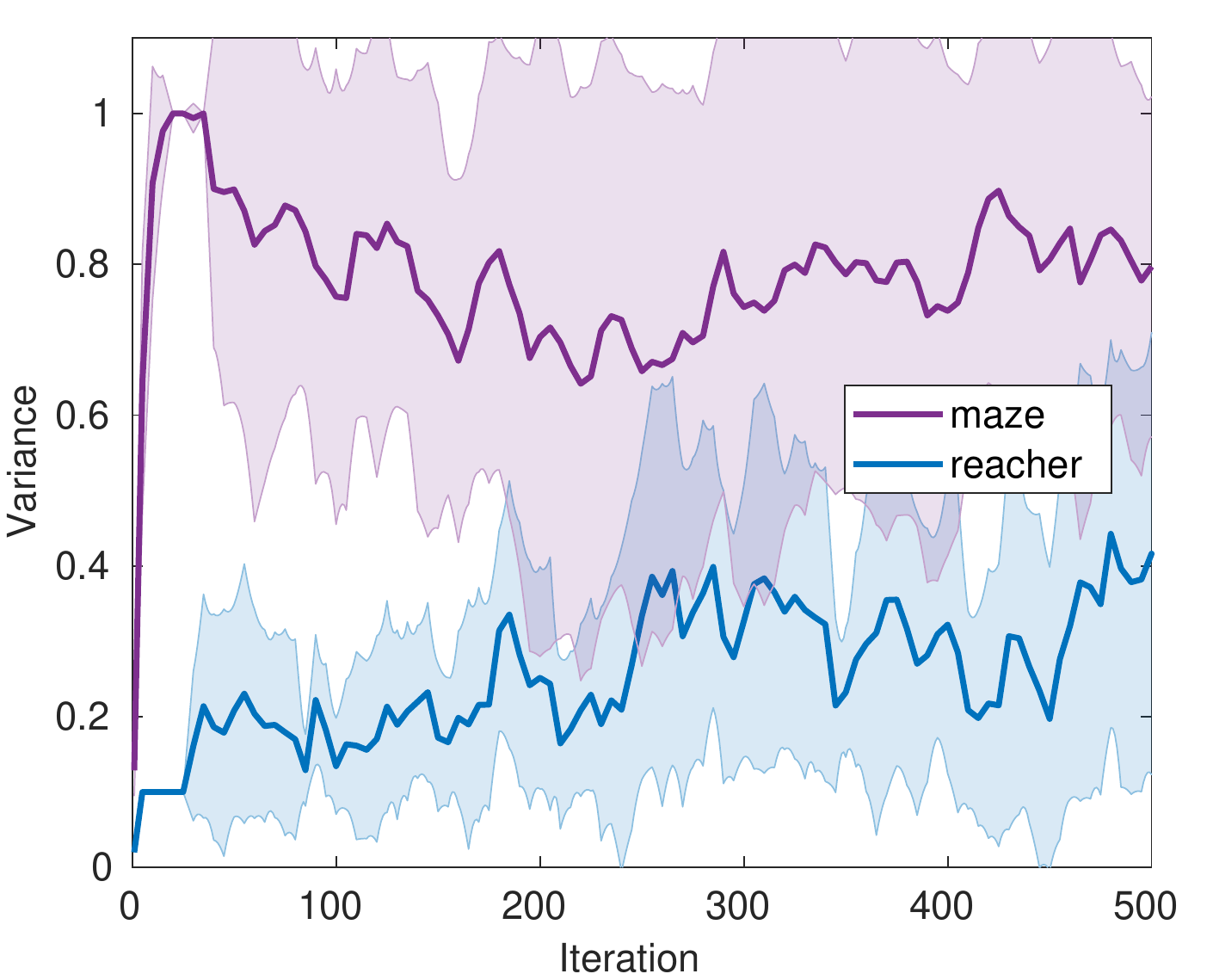}
\caption{Variance adaptation for different environments in the multi-goal scenario. The lines show the average and the shaded region the standard deviation over 6 executions.}
\label{variance_adapt}
\vspace{-10pt}
\end{figure}

Fig.~\ref{variance_adapt} shows the sampling variance evolution over training. 
Initially, our algorithm picks the largest and the smallest variance values for the Maze and the SparseReacher environments, respectively.
In the case of SparseReacher, it keeps a low variance at the beginning, since random initialization of the policy weights results in actions of large magnitude.
As the agent keeps learning, the exploration is gradually relaxed.
Our algorithm regulates the variance in such a way that allows the learner to maintain the proper exploration pace, resulting in steep learning curves.
We also find this idea connected to the approach proposed by~\cite{berthelot17} in the context of adversarial learning.
In our scenario, since there is no loss for the sampler, we apply the equilibrium principle through balancing the success ratio for the learner.

We also evaluated the single-goal variations of our environments, where the seed state  represents the only goal in each environment. 
In this scenario, variance adaptation showed similar benefits. 
On average the single-goal SparseReacher was learned 20--50\% faster with variance adaptation than with manual tuning of a constant variance.
For the Maze environment our algorithm is able to match the performance of the version with the constant sampler variance.





\section{Conclusions and Future work}
In this work we proposed a novel algorithm for learning a global policy capable of moving an agent in environments with any pair of start-goal transitions. 
Our algorithm is based on the idea of  region growing, and it is capable of automatic adjustment of the region expansion to achieve an appropriate pace of learning without extensive hyperparameter tuning.

To apply our approach to real robotic systems in the future, we plan to address the following shortcomings.
First, the algorithm could substantially benefit from parallel learning of a reversing policy, allowing it to return to some safe states within the current growing region. 
Second, the current version of our algorithm is sensitive to the choice of the seed state. 
For example, in experiments with the Maze environment, we observed that the success rate can be twice as good depending on the location of the seed state.
This phenomenon occurs because for some seeds the policy can avoid learning how to reach states in the hardest subspace of the environment, namely, the corridor.
In the future, we plan to address this problem by using a principle of skill chaining to learn a set of policies for different state regions.

\newpage
\appendix

\bibliographystyle{named}
\bibliography{bibliography}

\end{document}